\def\isarxiv{1}

\ifdefined\isarxiv
\documentclass[11pt]{article}

\usepackage[numbers]{natbib}

\else
\documentclass{article}
\usepackage{neurips_2023}
\fi

\usepackage{hyperref}
\usepackage{url}
\usepackage{graphicx,subcaption}
\usepackage{amsmath}
\usepackage{amsthm}
\usepackage{amssymb} 
\usepackage{caption}
\usepackage{subcaption}
\usepackage{wrapfig}
\usepackage{amsfonts}
\usepackage{color}
\usepackage{algpseudocode}
\usepackage{algorithm}

\newtheorem{theorem}{Theorem}[section]
\newtheorem{lemma}[theorem]{Lemma}
\newtheorem{definition}[theorem]{Definition}

\newtheorem{fact}[theorem]{Fact}

\newcommand{\wt}{\widetilde}

\newcommand{\R}{\mathbb{R}}

\renewcommand{\d}{\mathrm{d}}

\DeclareMathOperator{\poly}{poly}

\DeclareMathOperator{\nnz}{nnz}

\DeclareMathOperator{\diag}{diag}

\DeclareMathOperator{\reg}{reg}
\DeclareMathOperator{\cent}{cent}

\graphicspath{{./figs/}}

\ifdefined\isarxiv

\usepackage{tikz}
\usepackage{hyperref}  
\hypersetup{colorlinks=true,citecolor=blue,linkcolor=blue} 
\usetikzlibrary{arrows}
\usepackage[margin=1in]{geometry}

\else

\usepackage{microtype}
\usepackage{hyperref}
\definecolor{mydarkblue}{rgb}{0,0.08,0.45}
\hypersetup{colorlinks=true, citecolor=mydarkblue,linkcolor=mydarkblue}

\fi

\title{InfoPrompt: Information-Theoretic Soft Prompt Tuning for  Natural Language Understanding}

\begin{document}

\ifdefined\isarxiv

\date{}

\author{
Junda Wu\thanks{\texttt{jw6466@nyu.edu}. New York University.}
\and 
Tong Yu\thanks{\texttt{tyu@adobe.com}. Adobe Research.} 
\and 
Rui Wang\thanks{\texttt{rwr16@duke.edu}. Duke University.}
\and 
Zhao Song\thanks{\texttt{zsong@adobe.com}. Adobe Research.}
\and 
Ruiyi Zhang\thanks{\texttt{ruizhang@adobe.com}. Adobe Research.}
\and 
Handong Zhao\thanks{\texttt{hazhao@adobe.com}. Adobe Research.}
\and 
Chaochao Lu\thanks{\texttt{luccnju@gmail.com}. University of Cambridge.}
\and 
Shuai Li\thanks{\texttt{shuaili8@sjtu.edu.cn}. Shanghai Jiao Tong University.}
\and 
Ricardo Henao\thanks{\texttt{ricardo.henao@duke.edu}. Duke University.} 
}

\begin{titlepage}
  \maketitle
  \begin{abstract}
 
Soft prompt tuning achieves superior performances across a wide range of few-shot tasks. 
However, the performances of prompt tuning can be highly sensitive to the initialization of the prompts. 
We also empirically observe that conventional prompt tuning methods cannot encode and learn sufficient task-relevant information from prompt tokens.  
In this work, we develop an information-theoretic framework that formulates soft prompt tuning as maximizing mutual information between prompts and other model parameters (or encoded representations). This novel view helps us to 
develop a more efficient, accurate and robust soft prompt tuning method InfoPrompt. With this framework, we develop two novel mutual information based loss functions, to (i) discover proper prompt initialization for the downstream tasks and learn sufficient task-relevant information from prompt tokens and (ii) encourage the output representation from the pretrained language model to be more aware of the task-relevant information captured in the learnt prompt. 
Extensive experiments validate that InfoPrompt can significantly accelerate the convergence of the prompt tuning and outperform traditional prompt tuning methods. Finally, we provide a formal theoretical result for showing to show that gradient descent type algorithm can be used to train our mutual information loss.  

\end{abstract}
  \thispagestyle{empty}
\end{titlepage}
{
}
\else

\maketitle

\begin{abstract}
 
Soft prompt tuning achieves superior performances across a wide range of few-shot tasks. 
However, the performances of prompt tuning can be highly sensitive to the initialization of the prompts. 
We also empirically observe that conventional prompt tuning methods cannot encode and learn sufficient task-relevant information from prompt tokens.  
In this work, we develop an information-theoretic framework that formulates soft prompt tuning as maximizing mutual information between prompts and other model parameters (or encoded representations). This novel view helps us to 
develop a more efficient, accurate and robust soft prompt tuning method InfoPrompt. With this framework, we develop two novel mutual information based loss functions, to (i) discover proper prompt initialization for the downstream tasks and learn sufficient task-relevant information from prompt tokens and (ii) encourage the output representation from the pretrained language model to be more aware of the task-relevant information captured in the learnt prompt. 
Extensive experiments validate that InfoPrompt can significantly accelerate the convergence of the prompt tuning and outperform traditional prompt tuning methods. Finally, we provide a formal theoretical result for showing to show that gradient descent type algorithm can be used to train our mutual information loss.  
\end{abstract}

\fi

\section{Introduction}
Soft prompt tuning has shown great successes in a wide range of natural language processing tasks, especially in low-resource scenarios \cite{lester2021power,gu2022ppt,liu2022psp}. With a relatively small portion of tunable prompt parameters appended to the input of the context, the language model can be trained on the downstream tasks with the large scale pretrained parameters frozen. Compared with conventional fine tuning methods, prompt tuning requires less memory and computational resources to update these significantly smaller sized prompt parameters. In addition, in low-shot learning scenarios, prompt tuning can prevent the language model overfitting on the training data, and thus maintain the generalization ability of pretrained language models.

However, recent works reveal 
that it is non-trivial to find proper initialization of the prompt tokens.
Several works have investigated the effect of prompt initialization on the prompt tuning performances \cite{su2022transferability,wu2022adversarial} and showed that the performances of prompt tuning are highly sensitive to the prompt initialization.  
However, since the prompt initialization can be variant to different downstream tasks and pretrained language models, in low-resource tasks, we can hardly find very accurate knowledge to guide us to obtain proper initialization \cite{chen2022knowprompt}.

\begin{figure}[!ht]
     \centering
     \begin{subfigure}[b]{0.48\textwidth}
         \centering
         \includegraphics[width=\textwidth]{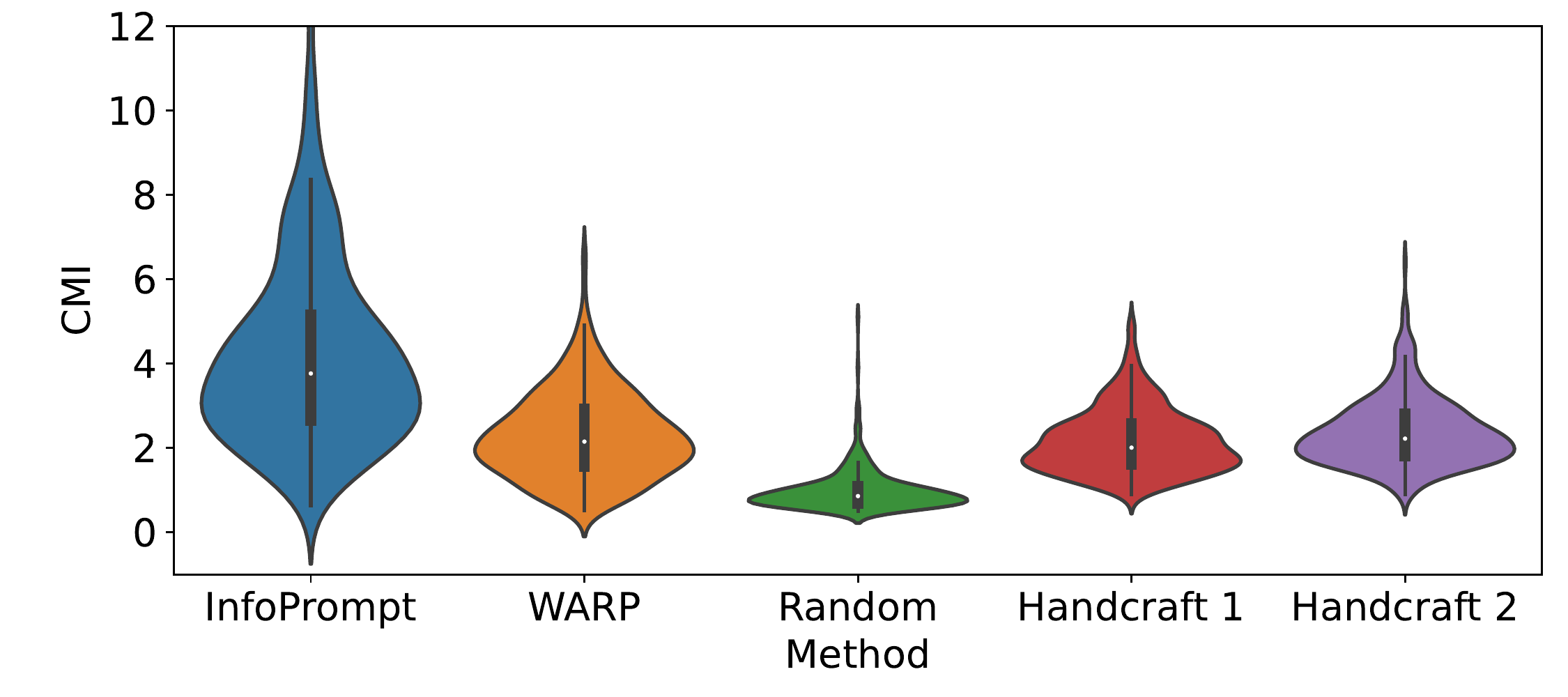}
         \caption{ 
         MRPC. Prompt handcraft 1 is `it is equivalent to' and prompt handcraft 2 is `has the same meaning'. 
         }
         \label{fig:MRPC-violin}
     \end{subfigure}
     \hfill
     \begin{subfigure}[b]{0.48\textwidth}
         \centering
         \includegraphics[width=\textwidth]{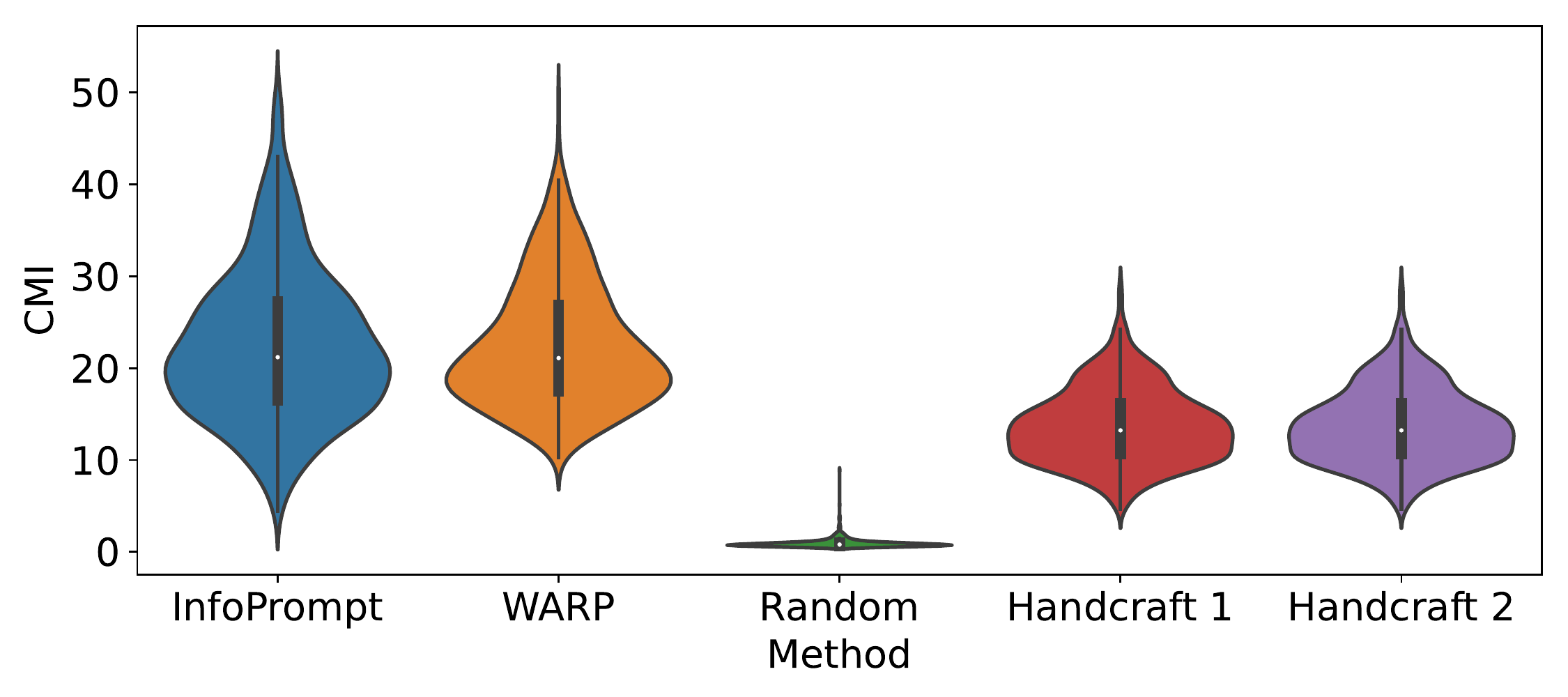}
         \caption{ 
         SST-2. Prompt handcraft 1 is `this is positive' and prompt handcraft 2 is `this is negative'.}
         \label{fig:SST-violin}
     \end{subfigure}
        \caption{Distributions of the CMI metrics of the prompts learned or handcrafted from different methods on MRPC and SST-2 \cite{wang2018glue}. By our InfoPrompt, the relevance between the prompt tokens and downstream tasks is the highest among all methods.}
        \label{fig:violin}
\end{figure}

In addition to the above limitations of soft prompt tuning in the initialization, we also empirically observe that conventional prompt tuning methods cannot effectively learn sufficient task-relevant information from prompt tokens.  
Specifically, conventional prompt tuning methods cannot locate prompts with task-relevant information encoded by the frozen language models conditioned on the input context. 
To understand the relevance between the prompt tokens and downstream tasks, we calculate the conditional mutual information (CMI) between the prompt tokens and the latent representation from the language model conditioned on the input context. We follow \cite{hambardzumyan2021warp} to determine the positions of prompt tokens inserted between the input sentence. Figure \ref{fig:violin} shows the distribution of CMI of the
prompts learned or handcrafted from different methods. The randomly sampled prompts have the lowest CMI. The prompts learned by a soft prompt tuning method, WARP \cite{hambardzumyan2021warp}, can have relatively better CMI than handcrafted prompts. By directly maximizing the CMI, our InfoPrompt (detailed in Section \ref{sec:method}) can encode more informative prompts.  
Without the guidance of task-relevant information, randomly exploring the prompts through the large continuous space of the prompt tokens can be inefficient, where a similar challenge is discussed in \cite{qin2021exp}. Some related results \cite{qin2021exp,gu2022ppt,schucher2022power} show that prompt tuning takes much larger numbers of epochs to converge comparing to conventional fine tuning methods.  
Our method can converge more easily than traditional soft prompt tuning method  
and we provide the theoretical guarantees of the convergence of the proposed losses which are trained using conventional gradient decent type algorithms.
 
In this paper, to address the above limitations, we develop an information-theoretic framework that formulates soft prompt tuning as maximizing mutual information between prompts and other model parameters (or encoded representations).  
With this framework, we develop InfoPrompt with two novel mutual information based loss functions. (i) To discover proper prompt initialization for the downstream tasks and learn sufficient task-relevant information from prompt tokens, we optimize the \emph{head loss} which maximizes the mutual information between the prompt and the language head. By optimizing this head loss, the prompt can effectively learn task-relevant information from the downstream tasks, since
the language head usually contains the information from the downstream tasks.
Besides, the mutual information loss can help to guide the learning of the prompt tokens in the early steps to tackle the initialization challenge, since language head parameters are more likely to learn the downstream task information more quickly.
(ii) To further encourage the output representation from the pretrained language model (\emph{i.e.}, encoded representation) to be more aware of the task-relevant information captured in the learnt prompt, we optimize the \emph{representation loss} which maximizes the conditional mutual information between the prompt tokens and the feature representations conditioned on the input context.

Our contributions are summarized as:
\begin{itemize}
    \item  
    We revisit the limitations of soft prompt tuning in the initialization and discover that existing prompt tuning methods cannot learn sufficient task-relevant information from prompt tokens.  
    We propose to jointly solve these limitations from an information-theoretic perspective. 
    \item We propose an information-theoretic prompt tuning framework InfoPrompt. We develop two novel loss functions to find proper prompt initialization and learn sufficient task-relevant information from prompt tokens,  
    without requiring any prior knowledge.
    \item Extensive experiments on 6 tasks and 3 datasets validate that, without any prior knowledge, InfoPrompt can significantly accelerate the convergence of the prompt tuning and outperform traditional prompt tuning methods.
    \item We provide a formal theoretical result to show that gradient descent type algorithm can be used to train our mutual information loss.
\end{itemize}
 
\section{Preliminary}

\subsection{Prompt Tuning} 
Soft prompt tuning has shown great successes in a wide range of NLP tasks \cite{lester2021power,gu2022ppt,liu2022psp}.  
Let $\Phi$ denote the encoder of a pretarined language model, \emph{e.g.}, Roberta-Large \cite{Liu2019RoBERTaAR}.
Assume $X=\{x_1, x_2, \cdots , x_n\}$ is a text sequence of length $n$ and $Y$ is its label. 
By prompting, we add extra information $P$ for the model to condition on during its prediction of $Y$.  
$P= \{p_1, \dots , p_{n_p}\}$  is a sequence of token embeddings.
$n_p$ is the number of prompt token. 
$p_i\in\mathbb R^D$, $i=1,\cdots,n_p$, is a embedding vector with dimension $D$ and $D$ is the embedding dimension of the pretrained language model. 
We first embed each token of $X$ into its corresponding token embedding from the pretrained language model.
$P$ is inserted into the the resulting embedding sequence of $X$ to form a template, which is further encoded by the pretrained encoder $\Phi$ into the representation space of the pretrained langauge model.
The template is detailed with the experiments.
Formally, we denote such a process as $Z=\Phi(P,X)$, with $Z$ being the encoded representation. 
The model prediction for $X$ is made on top of $Z$ with a trainable language modeling head $h_\theta$ (parameterized by $\theta$), \emph{s.t.}, the output $h_\theta(Z)$ is the distribution over all possible labels for classification. 
The whole model is generally trained via minimizing the following loss function,
\begin{equation}
    \mathcal L_{\mathrm{pred}}=\mathrm{cross}\_\mathrm{entropy}(h_\theta(Z),Y)
\end{equation}
 
\noindent The pretrained encoder $\Phi$ is frozen during training. 
Different from previous works \cite{lester2021power} where the prompts are learnable parameters, 
in our approach, the prompts are encoded from the input $X$, 
so that the prompts can encode task-relevant information from the training text $X$.
We will elaborate on how the prompts are encoded from input $X$ in Section \ref{sc: exp}.

\subsection{Mutual Information} \label{sc: MI}
Mutual information (MI) is a metric in information theory \cite{shannon1948mathematical,cover1999elements}, which measures the amount of shared information between two random variables. The mutual information between two random variables $A$ and $B$ is
\begin{equation}
    \begin{aligned}
        &\mathcal I(A;B)
        = &\mathbb{E}_{p(a,b)}\left[D_{KL}\left[p(a|b)\|p(b)\right]\right]. 
    \end{aligned}
\end{equation}

Inspired by \cite{sorensen2022information}, we use MI as the criterion for comparing prompts and other model parameters. MI has also been applied to measure the similarity between the masked token and the corresponding context in the pretraining of Multilingual Masked Language Modeling \cite{chi2021infoxlm}, the relevance between the document and a sentence in document summarization \cite{padmakumar2021unsupervised} and the source sentence and the target translated sentence in Neural Machine Translation (NMT) \cite{zhang2022conditional}.

\section{Our Method: InfoPrompt}
\label{sec:method}

As mentioned above, we want to the learnt prompts to be task-relevant. 
To achieve this, we notice that the language model head is trained with both the data representation and the classification labels, thus show be relevant to the task to be trained.
In encouraging the task-relevancy of the learnt prompts, we consider maximizing the mutual information between the prompt and the parameters of he language model head, denoted as $\theta$.
By maximizing such mutual information, the learnt prompt will be more correspondent with the training data with which the language model head is trained, thus captures more task-relevant information from training.
Further, in order for the pretrained language model to properly encode the task-relevant information in the prompt, we additionally maximize the mutual information between the prompt and the representation from the pretrained language model, so that the encoded representation can be aware of the task-relevant information captured by the prompt.  
In addition, we also provide theoretical guarantees of the convergence of those losses trained by conventional gradient decent type algorithms, which enables our method to converge more easily than traditional soft prompt tuning methods and avoid the problem of large training epochs. 
Below, we denote the negative mutual information between the prompt and parameters of the  language model head as the \textit{head loss}.
The negative mutual information between the prompt and the representation from the pretrained language model is denoted as the \textit{representation loss}.

 \subsection{The Head Loss}
\label{sc: head loss}
 
the head loss is the negative mutual information between the prompt $P$ and parameters $\theta$, \emph{i.e.}, $-I(P;\theta|X)$.
In maximizing $I(P;\theta|X)$, we follow \cite{ma2018noise} that approximate it with the following lower bound,

\begin{equation}
    \begin{aligned}
    &\mathcal I(P;\theta|X)\geq C + \mathcal L_{NCE}(P,\theta,X),
    \end{aligned}
\end{equation}

\noindent $C$ is a constant and $\mathcal L_{NCE}$ is a Noise Contrastive Estimation (NCE) of mutual information,
\begin{equation}
    \begin{aligned}
    &\mathcal L_{NCE} = \mathbb{E}\left[\log \frac{\exp(l(P,\theta,X))}{\sum_{k=1}^K \exp(l(P_k,\theta|X))}\right],
    \end{aligned}
\end{equation}
$\{P_k\}_{k=1}^K$ are the negative prompt samples for contrastive learning. In practice, we randomly sample $K-1$ tokens from the context as the negative samples, \emph{i.e.}, $\{P_k\}_{k=2}^K$ and $P_1=P$.

We model the score function $l(\theta,p)$ as a standard bilinear function with the learnable matrix $W_1$
\begin{equation}
    \begin{aligned}
    l(P,\theta|X)=P^\top W_1\theta.
    \end{aligned}
\end{equation}
where $\theta$ and $P$ are encoded from $X$.
$W_1$ is a trainable matrix.
Since the language model head is learnt on top of the output from the last layer of the pretrained language model, 
the learning of its parameters $\theta$ is easier than the learning of the prompt $P$ (the input layer of the pretrained language model).
Therefore, $\theta$ may capture more task-relevant information than $P$ in the early stage of training.
By maximizing the mutual information between $\theta$ and $P$, the task-relevant information can be transferred to $P$ in the initial training steps. 
In this way, $P$ can be more task-relevant especially in the early training stage.
Experiments in Section \ref{sc: ana} also show that our head loss, $\mathcal I(P;\theta|X)$, can facilitate the training of the initial training steps.

\subsection{The Representation Loss} \label{sc: info loss}
The representation loss, denoted as $-\mathcal I(P;Z|X)$, is defines as the negative of mutual information between the prompt $P$ and the encoded representation from the pretrained language model, \emph{i.e.}, $Z=\Phi(P,X)$. Similar to the head loss, we approximate the representation loss with its lower bound,
\begin{equation}
    \begin{aligned}
    &\mathcal I(P;Z|X)\geq \log(N) + \mathcal L_{NCE}(P,Z|X),
    \end{aligned}
\end{equation}
\noindent and,
\begin{equation}
    \begin{aligned}
    &\mathcal L_{\mathrm{NCE}} = \mathbb{E}\left[\log \frac{\exp(l(P,Z|X))}{\sum_{k=1}^K \exp(l(P,Z_k|X))}\right],
    \end{aligned}
\end{equation}
$\{Z_k\}_{k=1}^K$ are the negative samples.
Here, we overload the notations of InfoNCE loss $\mathcal L_{NCE}$ and score function $l$ for conciseness. 
Let $W_2$ be a trainable matrix, the function $l$ for the representation loss is defined by,
\begin{equation}
    \begin{aligned}
    l(P,Z|X)=P^\top W_2Z.
    \end{aligned}
\end{equation}
We use variational inference methods \cite{houthooft2016vime} to recover the latent distribution of $Z$.
Specifically, we assume that the latent distribution is $N(\mu,\sigma)$, where $N(\mu,\sigma)$ is the normal distribution with mean $\mu$ and diagonal covariance matrix $\sigma$. 
We model $\mu$ and $\sigma$ via,
\begin{equation}
    \begin{aligned}
    \mu = f_\mu(Z), \sigma = f_\sigma(Z).
    \end{aligned}
\end{equation}
$f_\mu$ and $f_\mu$ are trainable fully connected layers.
Since the negative samples of $Z$, \emph{i.e.}, $\{Z_k\}_{k=1}^K$, should not be paired with $P$, we assume the $\{Z_k\}_{k=1}^K$ are drawn from $N(\mu^\prime, \sigma^\prime)$, \emph{s.t.},
\begin{equation}
    \begin{aligned}
    \mu^\prime = f_\mu(Z^\prime), \sigma^\prime = f_\sigma(Z^\prime).
    \end{aligned}
\end{equation}
In contrast to $Z=\Phi(P,X)$, we have $Z^\prime=\Phi(X)$ so that $\{Z_k\}_{k=1}^K$ are not paired with $P$.
By maximizing the representation loss $I(P;Z|X)$, we encourage the encoded representation $Z$ to be more aware of the prompt $P$, so that the task-relevant information in $P$ can be properly encoded by the pretrained language model in producing $Z$.

\subsection{Overall Objective}
 
We minimize the following objective in prompt tuning:
\begin{equation} \label{eq:1}
    \mathcal L 
    =\mathcal L_{\mathrm{pred}} -\beta \cdot \mathcal I(P;Z|X)-\gamma \cdot \mathcal I(P;\theta|X) 
    .
\end{equation}
 
We denote $\mathcal L_{\mathrm{pred}}$ as the task loss.
$\beta$ and $\gamma$ are balancing parameters for the proposed representation loss and head loss, respectively.
We denote our approach as \textit{InfoPrompt}.

\subsection{Theoretical Guarantees}
 
We state our main theoretical result as follows. Due to the space limit, we delay the proof into Appendix.
\begin{theorem}\label{thm:main:informal}
Given the Loss function ${\cal L}$ (Eq.~\eqref{eq:1}) and conditions specified in Appendix C.1 and D, using gradient descent type of greedy algorithm, we can find the optimal solution of that loss function.
\end{theorem}

We provide theoretical guarantees of the convergence of those losses trained by conventional gradient decent type algorithms, which enables our method to converge more easily than traditional soft prompt tuning methods and avoid the problem of large training epochs. 

\section{Experiments} \label{sc: exp}

\subsection{Datasets}
 
We conduct experiments with datasets of sequence classification from the GLUE benchmark \cite{wang2018glue}, along with those of relation extraction tasks and NER tasks.
We choose four seqience classification tasks from the GLUE benchmark: RTE (Recognizing Textual Entailment, \cite{bentivogli2009fifth}), MRPC (Microsoft Research Paraphrase Corpus, \cite{dolan2005automatically}), CoLA (Corpus of Linguistic Acceptability, \cite{warstadt2019neural}) and SST-2 (Sentence Sentiment Treebank, \cite{socher2013recursive}). 
We choose these tasks because their datasets are in smaller sizes and prompt tuning is comparably more effective in low-resource setting \cite{hambardzumyan2021warp, li2021prefix}.
For the task of relation extraction, we evaluate our method on ACE2005 corpus and Semeval-2010 datasets \cite{hendrickx2019semeval}.  
We also use the ACE2005 corpus for the task of NER.
Note that the entity spans for NER have been given ACE2005.
Unlike the standard NER model that learns to predict the entity span and entity type simultaneously from the raw text sequence, our model only predicts the entity type based on the given entity span.
We follow the same data splitting strategy for ACE2005 corpus as the previous work \cite{yang2016joint,nguyen2016joint}. For the Semeval-2010 tasks, we follow the official data partition \cite{hendrickx2019semeval}.

\subsection{Experiment Settings}

We follow the resource constrained scenario in \cite{hambardzumyan2021warp} that train each task with only 64 or 256 samples. 
We experiment with $n_p=1$ and $n_p=4$ prompt tokens for each task.
The prompt tokens are inserted into the template for each task.
Similar to \cite{hambardzumyan2021warp}, we adopt the RoBERTa-large model as our pretrained encoder.
We freeze pretrained parameters are  only trained with parameters of the prompt head and prompt tokens. 
During training, we empirically set $\beta=0.1$ and $\gamma=0.05$.
The number of negative of negative samples $K=32$.
The learning rate is $1e-3$ and the batch size is 8. 
For each task, we report the results after 30 epoches, averaged over 5 random seeds. 
To encode the prompt $P=[p_1,\cdots,p_{n_p}]$ from $X$, we first encode $X$ into $P^\prime\in\mathbb R^D$ via $P^\prime=\Phi(X)$. 
For each $p_i\in\mathbb R^D$, we have $p_i=W_i^{\mathrm{up}}W_i^{\mathrm{down}}P^\prime$, $W_i^{\mathrm{up}}\in\mathbb R^{D\times 64}$, $W_i^{\mathrm{down}}\in\mathbb R^{64\times D}$.

For the tasks of sequence classification and relation extraction, we follow the template of \cite{hambardzumyan2021warp} that contains a $\mathsf{[mask]}$ token.
The representation $Z$ is the $\mathsf{[mask]}$ token from the last layer of the  Roberta-Large encoder.
For the tasks of named entity recognition, we have the $\mathsf{[mask]}$ token before the given entity span, with the rest being the same as  for sequence classification.

\begin{table*}[htp]
\caption{Results on Sequence Classification. 
}
\label{tab:glue}
\resizebox{1.0\textwidth}{!}{%
\begin{tabular}{l|cc|cc|cc|cc|c}
\hline
                         & CoLa ($n_p=1$) & CoLa ($n_p=4$) & RTE ($n_p=1$) & RTE ($n_p=4$) & MRPC ($n_p=1$) & MRPC ($n_p=4$) & SST2 ($n_p=1$) & SST2($n_p=4$) & Average \\ \hline
Finetuning        & \multicolumn{2}{c|}{0.6131} & \multicolumn{2}{c|}{0.7798} & \multicolumn{2}{c|}{0.8873} & \multicolumn{2}{c|}{0.9427} & 0.8057 \\
Adapter           & \multicolumn{2}{c|}{0.5552} & \multicolumn{2}{c|}{0.5776} & \multicolumn{2}{c|}{0.6814} & \multicolumn{2}{c|}{0.9472} & 0.6904 \\ \hline
WARP  \cite{hambardzumyan2021warp}                     &  0.5282  &  0.5911  &  0.6282  &  0.6426  &  0.8039  &  0.8186  &  0.9507  &  0.9587  &  0.7403  \\
IDPG \cite{wu2022idpg}                      &  0.5556  &  0.5646  &  0.6282  &  0.6534  &  0.7941  &  0.8039  &  0.9587  &  0.9587  &  0.7396  \\
InfoPrompt                 &  0.5631  &  0.6018  &  0.6751  &  0.6968  &  0.8039  &  0.8137  &  0.9576  &  0.9599  &  0.7590  \\
~~ $\gamma=0$                &  0.5699  &  0.5853  &  0.6751  &  0.6787  &  0.7941  &  0.8137  &  0.9495  &  0.9587  &  0.7531  \\
~~ $\beta=0$                &  0.5546  &   0.5579     &  0.6065  &  0.6318  &  0.7892  &  0.7966  &  0.9472  &  0.9610  & 0.7306   \\
~~ $\gamma=0,\beta=0$        &  0.5032  &  0.5732  &  0.6173  &  0.6029  &  0.7917  &  0.7672  &  0.9495  &  0.9564  &  0.7202  \\ \hline
\end{tabular}%
}
\end{table*}

\begin{table*}[htp]
\centering
\caption{Results on Relation Extraction and NER.}
\label{tab:re-ner}
\resizebox{1.00\textwidth}{!}{%
\begin{tabular}{l|cc|cc|cc|c}
\hline
                          & RE ($n_p=1$) & RE ($n_p=4$) & NER($n_p=1$) & NER($n_p=4$) & SemEval ($n_p=1$) & SemEval ($n_p=4$) & Average \\ \hline
Finetuning                & \multicolumn{2}{c|}{0.8119} & \multicolumn{2}{c|}{0.9054} &      \multicolumn{2}{c|}{0.8506} &  0.8560      \\
Adapter                 & \multicolumn{2}{c|}{0.5073} & \multicolumn{2}{c|}{0.8329} &      \multicolumn{2}{c|}{0.6570}  &  0.6657      \\ \hline
WARP  \cite{hambardzumyan2021warp}                   &  0.6384  &  0.6596  &  0.8174  &  0.8607  &  0.6702  &  0.7284  &  0.7291  \\
IDPG      \cite{wu2022idpg}              &  0.6079  &  0.6132  &  0.8360  &  0.8931  &  0.6408  &  0.6776  &  0.7114  \\
InfoPrompt                  &  0.6914  &  0.7616  &  0.8526  &  0.8962  &  0.7563  &  0.7917  &  0.7916  \\
~~ $\gamma=0$               &  0.6914  &  0.7285  &  0.8452  &  0.8635  &  0.7471  &  0.7865  &  0.7770  \\
~~ $\beta=0$                &  0.6967  &  0.7470  &  0.8351  &  0.8698  &  0.7449  &  0.7538  &  0.7746  \\
~~ $\gamma=0,\beta=0$       &  0.5364  &  0.7285  &  0.8512  &  0.8661  &  0.7490  &  0.7799  &  0.7519  \\ \hline

\end{tabular}
}
\end{table*}

\subsection{Baselines and Ablations}
As mentioned above, our method with \eqref{eq:1} in denoted as InfoPrompt. In the experiments, we compare our method with the following baselines:
\begin{itemize} 
    \item Fine Tuning: We fine tune all the parameters from the pretrained encoder on each task. Fine Tuning is included as the upper bound for the model performance, since it is more computational expensive compared with only training the prompt parameters.
    \item Adapter: Similar to prompt tuning, this is also a way of parameter-efficient training for pretrained language models. Specifically, instead of adding the prompt tokens in the input, we add adaptors after the last linear layer in each transformer layer.
    \item WARP \cite{hambardzumyan2021warp}: Different from our approach, the prompt tokens of WARP are not generated from the input sequence. the prompt tokens are insert into the input sequence. During training, the pretrained encoder is frozen and only the prompt tokens are trainable.
    \item IDPG \cite{wu2022idpg}: Similar to our approach, the prompt tokens are generated from the input sequence. The pretrained encoder is frozen and the prompt generator is trainable. 
\end{itemize}
In evaluating the effectiveness of our proposed loss functions, we consider the following two ablations:
\begin{itemize}
    \item $\gamma=0$: We disable the head loss during training via $\gamma=0$, while keeping $\beta=0.05$.
    \item $\beta=0$: We disable the head loss during training via $\gamma=0$, while keeping $\gamma=0.1$.
    \item $\beta=\gamma=0$: We disable both the losses. The prompt parameters are trained with $L_{\mathrm{pred}}$.
\end{itemize}

\section{Experimental Results}

\subsection{Training with the Full dataset}

Table \ref{tab:glue} and \ref{tab:re-ner} show that results of training with the full dataset for each task.
We can observe that the results with our InfoPrompt is generally higher than the other parameters-efficient baselines that freeze the pretrained Roberta-Large parameters (\emph{e.g.}, WARP and Adapter).
Fine Tuning generally have better performance than the other approaches.
This is because it allows training with all the model parameters, which is at the expense of more computation cost during training. 
As mentioned above Fine Tuning is intended to be included as the upper bound for performance.
Moreover, we can find the performance with $\gamma=0$ and $\beta=0$ is lower than that of InfoPrompt, indicating shows that it is beneficial to learn task-relevant prompt tokens with the propose head loss and representation loss.
Further, the performance gap between $\gamma=0$/$\beta=0$ and $\beta=\gamma=0$ shows that the proposed functions are effective when added to naive prompt tuning, \emph{i.e.}, with only $\mathcal L_{pred}$.

\subsection{Training with the Few-Shot Datasets}

\begin{table*}[htp]
\caption{Few-shot results on Sequence Classification. We experiment with 64 and 256 samples for each task. The number of prompt is fixed to $n_p=4$ for all soft prompt tuning methods.}
\label{tab:glue-few}
\resizebox{\textwidth}{!}{%
\begin{tabular}{l|cc|cc|cc|cc|c}
\hline
                          & CoLa (64) & CoLa (256) & RTE (64) & RTE (256) & MRPC (64) & MRPC(256) & SST2 (64) & SST2(256) & Average \\ \hline
Finetuning                    &  0.1746  &  0.4086  &  0.4801  &  0.6787  &  0.7107  &  0.7819  &  0.8027  &  0.8853  &  0.6153  \\
Adapter                       &  0.0627  &  0.2486  &  0.5487  &  0.5668  &  0.5931  &  0.625  &  0.4908  &  0.664  &  0.4750  \\  \hline
WARP   \cite{hambardzumyan2021warp}                   &  0.0749  &  0.0785  &  0.5596  &  0.5812  &  0.7083  &  0.7083  &  0.5872  &  0.7638  &  0.5077  \\
InfoPrompt                    &  0.1567  &  0.1750  &  0.6137  &  0.6580  &  0.7059  &  0.7377  &  0.6697  &  0.7305  &  0.5559  \\
~~ $\gamma=0$                 &  0.1479  &  0.1447  &  0.5776  &  0.6318  &  0.6936  &  0.7328  &  0.664  &  0.7294  &  0.5402  \\
~~ $\beta=0$                  &  0.1372  &  0.1433  &  0.5812  &  0.5957  &  0.6838  &  0.7132  &  0.5631  &  0.656  &  0.5092  \\
~~ $\gamma=0, \beta=0$        &  0.0919  &  0.1397  &  0.5668  &  0.5523  &  0.6985  &  0.7108  &  0.5505  &  0.6296  &  0.4925  \\ \hline
\end{tabular}%
}
\end{table*}

\begin{table*}[htp]
\centering
\caption{Few-shot results on Reltion Extraction and NER. We experiment with 64 and 256 samples for each task. The number of prompt is fixed to $n_p=4$ for all soft prompt tuning methods.}
\label{tab:ner-re-few}
\resizebox{0.80\textwidth}{!}{%
\begin{tabular}{l|cc|cc|cc|cc|c}
\hline
                            &   RE (64)   &   RE (256)   &   NER (64)   &   NER (256)   &   SemEval(64)   &   SemEval(256)   &  Average  \\ \hline
Finetuning                    &  0.1285  &  0.4013  &  0.3033  &  0.4358  &  0.2223  &  0.4829  &  0.3290  \\
Adapter                       &  0.1086  &  0.1815  &  0.2345  &  0.2437  &  0.1211  &  0.177  &  0.1777  \\ \hline
WARP     \cite{hambardzumyan2021warp}                 &  0.1404  &  0.2556  &  0.3082  &  0.4369  &  0.1708  &  0.3684  &  0.2801  \\
InfoPrompt                    &  0.2119  &  0.2993  &  0.3331  &  0.4739  &  0.2113  &  0.4034  &  0.3222  \\
~~ $\gamma=0$                 &  0.2026  &  0.2834  &  0.3225  &  0.4776  &  0.2153  &  0.3739  &  0.3126  \\
~~ $\beta=0$                  &  0.2013  &  0.2874  &  0.3208  &  0.4615  &  0.2072  &  0.3629  &  0.3069  \\
~~ $\gamma=0,\beta=0$              &  0.1974  &  0.2728  &  0.3142  &  0.4662  &  0.2278  &  0.3276  &  0.3010  \\ \hline

\end{tabular}%
}
\end{table*}

The results for training with few-shot datasets are listed in Table \ref{tab:glue-few} and \ref{tab:ner-re-few}.
Compared with training with the full dataset (Table \ref{tab:glue} and \ref{tab:re-ner}), we can find that the performance gap between our propose InfoPrompt and the baselines is generally larger in the few-shot setting.
Unlike the full datasets, the few-shot datasets contains much less information regarding the task to be learnt.
As the result, the prompts learnt with solely the task loss (\emph{e.g.}, with WARP or $\beta=\gamma=0$) may easily overfit to the task-irrelvant information given few-shot datasets.
In such a scenario, it would be important to explicitly encourage the learnt prompt to be task-relevant, \emph{i.e.}, via our proposed loss functions based on mutual information maximization.
This explains why InfoPrompt yields larger performance gain when trained with few-shot datasets.
Similar to training with the full datasets, the performance gains of InfoPrompt  compared with InfoPrompt ($\gamma=0$) and InfoPrompt ($\beta=0$) show the effectuveness of our proposed loss functions in the few-shot scenario.

\section{Analysis} \label{sc: ana}

\begin{figure}[!ht]
    \centering
    \subfloat[The landscape of task loss. $\beta=0.1$]{\includegraphics[width=0.45\textwidth]{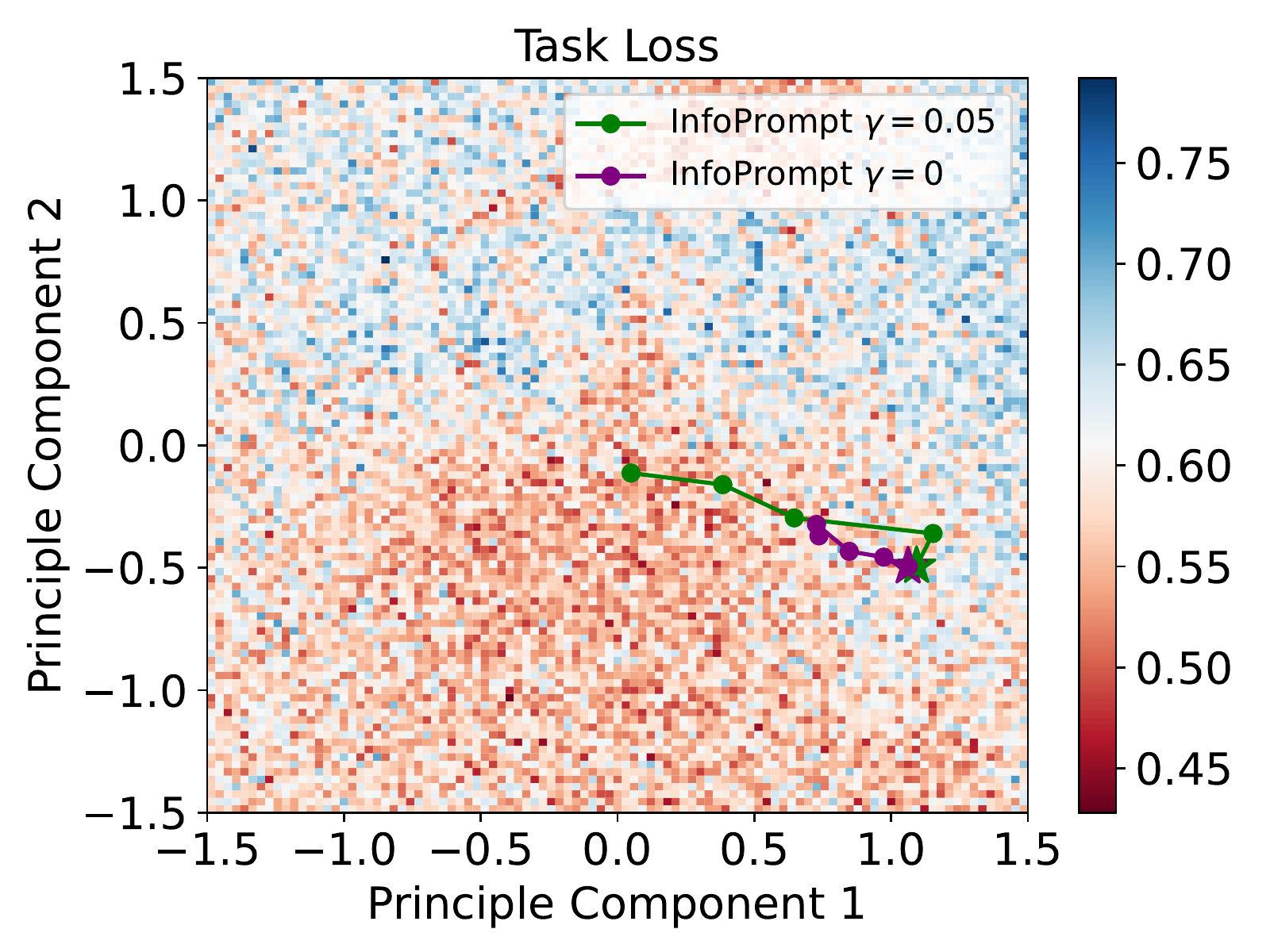}\label{fig: task loss}}
    \hspace{2mm}
    \subfloat[The landscape of representation loss. $\beta=0.1$]{\includegraphics[width=0.45\textwidth]{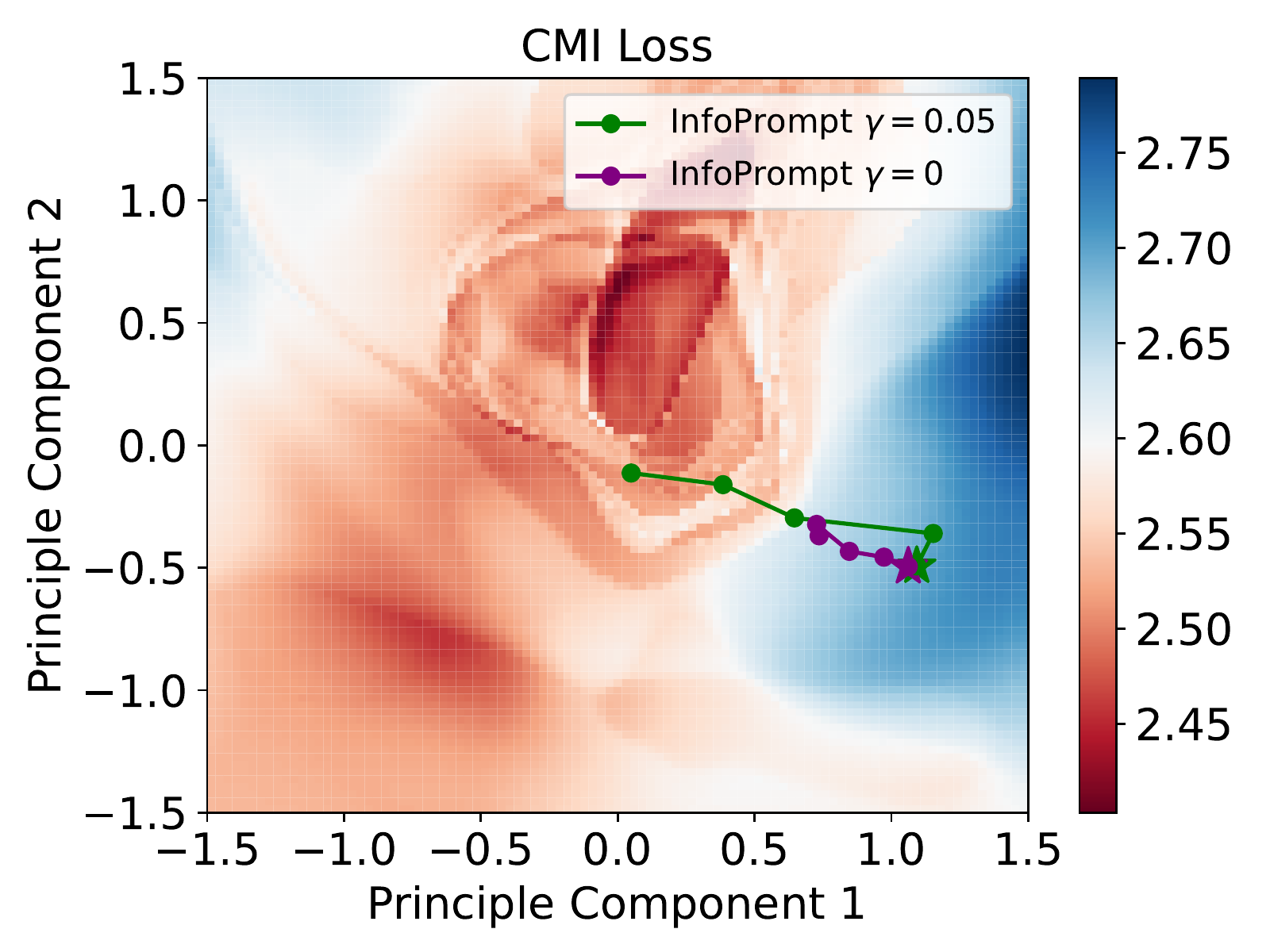}\label{fig: info loss}}
    \caption{The landscapes of the loss functions in the parameter space of the prompt tokens. The landscapes illustrates how the values of loss functions varies with the input prompt tokens. The trajectory shows the first 500 steps during training for InfoPrompt with $\gamma=0.05$  or $\gamma=0$.}
\end{figure}

\subsection{Loss Landscapes} \label{sc: land}
To provide a more comprehensive understanding on the effectiveness of our proposed loss functions, we plot the landscapes of the loss functions in the parameter space of the prompt tokens.
The landscapes illustrates how the values of loss functions varies with the input prompt tokens. 
Since the prompt tokens are high-dimensional vertors, \emph{i.e.}, each token has the dimension of 1024 for RoBerta-Large, we visualize their associated loss values via projecting the prompt tokens into a 2D subspace.
Specifically, we follow previous work of token embedding analysis \cite{cai2020isotropy} that project the prompt tokens into the top-2 principled component computed from the pretrained token embeddings of Roberta-Large. 
We only insert one prompt token into the input sequence during visualization.

Taking the task of MRPC as an example, we plot the 2D landscapes of the task loss and the representation loss in Figure \ref{fig: task loss} and \ref{fig: info loss}, respectively.
All the two figures are ploted with the same scale, \emph{i.e.}, with the same values of the prompt token.
The axis values are the offset from the mean of the pretrained Roberta-Large token embeddings. 
The loss values shown in the figures are the average of 20 random samples from MRPC.
In Figure \ref{fig: task loss}, we can find that there are a lot of local minimum in the landscapes of the task loss.
This  is consistent with the observations of the previous works \cite{gu2022ppt, su2022transferability} that prompt tuning is difficult to be optimized with and sensitive to initialization, \emph{e.g.}, the optimization can get easily overfit to a local minimum without proper initialization.
From Figure \ref{fig: info loss}, we can observe that the values of our proposed info loss is much smoother compared with the task loss in Figure \ref{fig: task loss}. 
With smoother landscapes, the optimization with our proposed loss functions can be more stable (also shown in Section \ref{sc: lc}), \emph{i.e.}. less likely being trapped in a local minimum  and also guaranteed to converge by our theoretical results (see Theorem~\ref{thm:main:informal}). 
Additionally, we plot the trajectory of the first 500 steps during training for InfoPrompt ($\gamma=0.05$) (green) and $\gamma=0$ (purple) in Figure \ref{fig: task loss} and \ref{fig: info loss}.
The stars in the plot indicate the initial value of the prompt before training.
We find that training with $\gamma=0.05$ can render a larger descent for both the task loss and representation loss, compared with $\gamma=0$.
As analyse in Section \ref{sc: head loss}, the language head is easier to be learnt than the prompt. 
As the result, parameters of the language head may contain more task-relevant information during the earlier stage of training.
By maximization the mutual information between the head parameter and prompt via the proposed head loss (weighted by $\gamma$), we encourage the learnt prompt to capture more task-relevant information in the initial training steps, thus resulting $\gamma=0.05$ to have a larger descent than  $\gamma=0$ in the trajectories shown in Figure \ref{fig: task loss} and \ref{fig: info loss}.  
Note that our propose two loss functions are unsupervised and do not required additional labels.

\subsection{Learning Curve} \label{sc: lc}

\begin{figure}[!ht]
    \centering
    \subfloat[NER ACE learning curve.]{\includegraphics[width=0.45\textwidth]{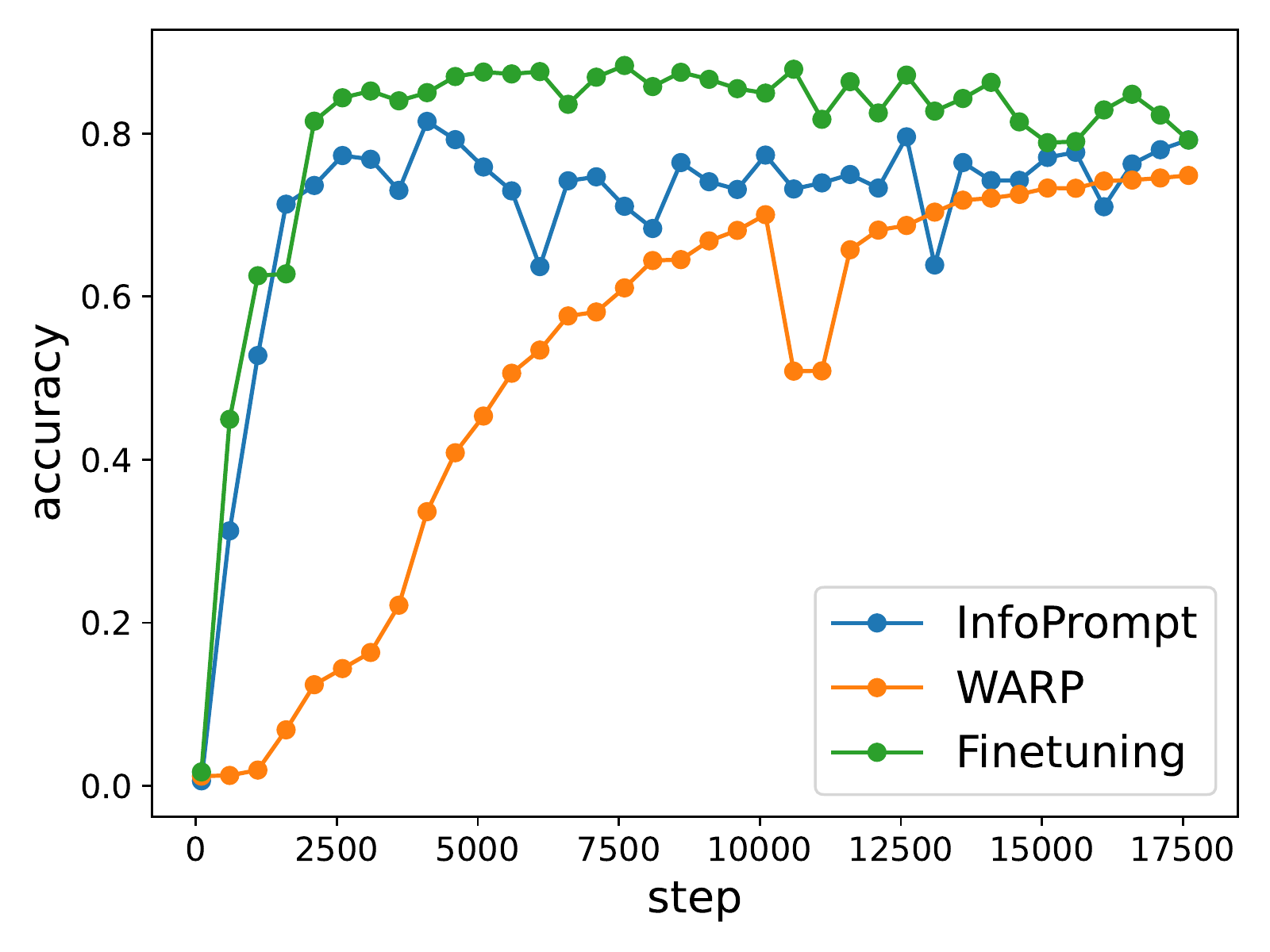}\label{fig: lc-ner}}
    \vspace{4mm}
    \subfloat[SST-2 learning curve.]{\includegraphics[width=0.45\textwidth]{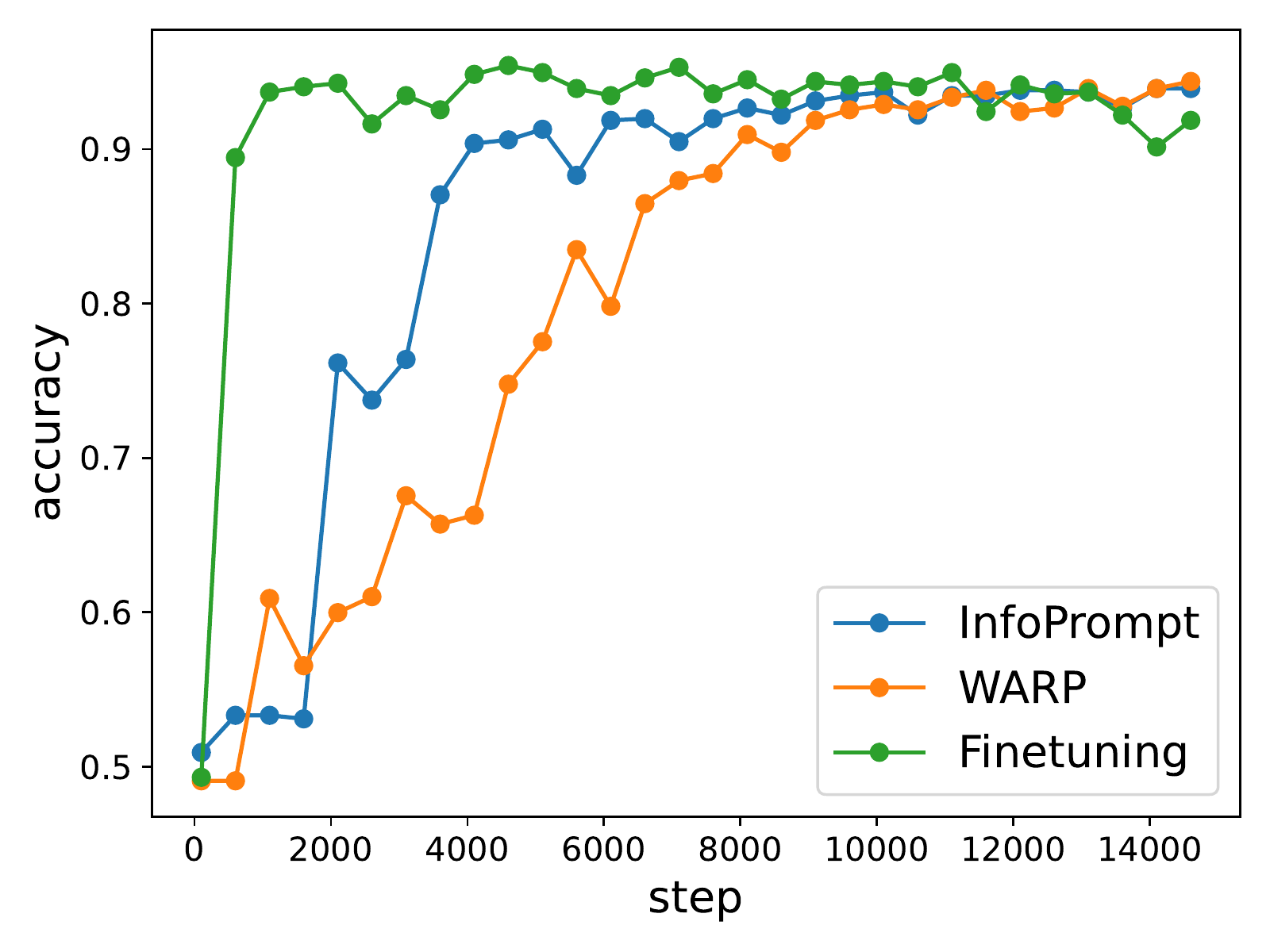}\label{fig: lc-sst}}
    \caption{The learning curves for the task of NER and SST-2. The training of our our InfoPrompt is more stabilized and converges faster, compared with WARP.} 
\end{figure}

We plot the training curve for the task of NER and SST-2 in Figure \ref{fig: lc-ner} and \ref{fig: lc-sst}, respectively.
Unlike WARP \cite{hambardzumyan2021warp} and Fine Tuning that train with solely the task loss $L_{\mathrm{pred}}$, our InfoPrompt also train with the representation loss and head loss. 
We can observe that the training of our our InfoPrompt is more stabilized and converges faster, compared with WARP. 
This can be explained from the landscape plots in Section \ref{sc: land}.
Since the landscape of the task loss is not smooth (Figure \ref{fig: task loss}), the training curve of WARP may exhibit significant perturbation when the optimization overfits to a local minimum, \emph{e.g.}, the 10000th in Figure \ref{fig: lc-ner}. 
Comparably, our proposed InfoPrompt can smooth the optimization landscape, thus stabilize the training and result in faster convergence, which is guaranteed by our theoretical results.
We observe that Fine Tuning generally converges faster and end up with a higher accuracy than InfoPrompt.
This is because Fine Tuning, which trains with all the model parameters, have much larger model capacity during training than prompt tuning (InfoPrompt and WARP).
Such results for Fine Tuning is at the expense of larger computation cost, \emph{i.e.}, we need to calculate the gradient for all the model parameters (354M) instead of only the prompt parameters $P$ (1.3M).

\section{Related Work}

\subsection{Soft Prompt Tuning}
Soft prompt tuning becomes a new paradigm in NLP. Based on some large pretrained models (e.g., BERT \cite{devlin-etal-2019-bert}, RoBERTa \cite{Liu2019RoBERTaAR}), a relatively small number of trainable parameters can be added to the input, while the parameters of backbones are fixed. Many works have demonstrated the effectiveness of soft prompt tuning in a wide range of NLP downstream tasks \cite{lester2021power, hambardzumyan2021warp, qin2021learning}, especially in low-resource regions \cite{schucher2022power, liu2022psp, gu2022ppt}. Some recent works also found the transferable power of soft prompts across domains \cite{wu2022adversarial, su2022transferability, vu2022spot}, across language models \cite{su2022transferability} and for zero-shot generalization \cite{ye2022retrieval}. To further enhance the efficiency of soft prompt parameters and enable better generalization abilities, many works consider multi-task learning \cite{asai2022attempt, ding2022prompt, vu2022spot, he2022hyperprompt}, or multilingual \cite{chen2022multilingual, huang2022zero}.
Some works also try to explore the prompt with prior knowledge encoded \cite{hu2021knowledgeable,he2022unified,chen2022knowprompt}.
While most of the initial attempts of soft prompt tuning are not context-aware, some recent works suggest that the soft prompt tokens should be conditioned on the input context. 
Hyperprompt \cite{he2022hyperprompt} proposes a hyper-network structure to generate prompt tokens based on task indexes. \cite{wu2022idpg} and \cite{bhardwaj2022vector} suggest some context-aware soft prompt generation methods. 
\cite{liu2022structured} proposes a structured soft prompt tuning method.

\subsection{Information-theoretic Methods in NLP}
Information-theoretic methods are widely used in many NLP tasks \cite{ju2021leveraging,west2019bottlesum,steinborn2022information,ji2022increasing,mathewson2019shaping}. \cite{west2019bottlesum} and \cite{ju2021leveraging} propose information-theoretic methods for text memorization. \cite{mathewson2019shaping} suggests an information-theoretic method for dialogue. \cite{ji2022increasing} views the multimodal NMT problem in an information-theoretic point of view. For model pretraining, \cite{wang2020infobert} proposes Infobert to improve the robustness of the BERT model. INFOXLM \cite{chi2021infoxlm} proposes a cross-lingual language model based on an information-theoretic framework. For model fine-tuning, \cite{mahabadi2021variational} proposes an information bottleneck model method for low-resource fine-tuning. \cite{sorensen2022information} proposes an information-theoretic method to engineer discreet prompts.

\subsection{Theoretical Attention Computation}

Softmax is one of the major unit in the attention scheme of most recent NLP large language models. Computing the attention matrix faster is a practically interesting question \cite{cklm19,linformer20,reformer20}. Recently, a number of theoretical work have tried to study the softmax/attention unit from theoretical perspective. The softmax attention computation can be formally defined as follows: suppose we are given three matrices $Q \in \R^{n \times k}$ (the query), $K \in \R^{n \times k}$ (the key), and $V \in \R^{n \times k}$ (the value), the goal is to compute $\mathsf{Att}(Q,K,V) = D^{-1} \exp(QK^\top) V$ where the diagonal matrix $D$ is $\mathrm{diag}( \exp(QK^\top) {\bf 1}_n )$. Here $K^\top$ denote the transpose of matrix $K$. The work of \cite{zhdk23,as23} consider the static setting, and the work of \cite{bsz23} considers the dynamic setting.  \cite{as23} proposed a tight algorithm for computing $\mathsf{Att}$ and provided a lower bound result based on the strong exponential time hypothesis. The work \cite{bsz23} shows a tight positive result and a negative result. In \cite{bsz23}, they provide an upper bound via lazy update techniques \cite{cls19}. In \cite{bsz23}, they also present a lower bound result which is based on the Hinted MV conjecture \cite{bns19}. The work of \cite{dms23} proposes two sparsification algorithm to compute attention matrix when the feature dimension $\gg$ the length of sentence. \cite{gsy23_dp} shows how to provide a differentially private algorithm for computing attention matrix under differential privacy framework \cite{dmns06,dkm+06}.

\section{Conclusion}

We revisit the limitations of soft prompt tuning in the initialization. We also empirically discover that conventional prompt tuning methods cannot learn sufficient task-relevant information from prompt tokens. 
We tackle these limitations from an information-theoretic perspective and propose an information-theoretic prompt tuning method InfoPrompt. Two novel loss functions are designed to (i) discover proper prompt initialization for the downstream tasks and learn sufficient task-relevant information from prompt tokens and (ii) encourage the output representation from the pretrained language model to be more aware of the task-relevant information captured in the learnt prompt.
With extensive experiments, without any prior expert knowledge, InfoPrompt can significantly accelerate the convergence of the prompt tuning and achieve more accurate and robust performances than traditional prompt tuning methods.

\section{Limitation}

Our proposed InfoPropmt is an approach that targets parameter-efficient training, \emph{i.e.}, reducing the trainable parameters via freezing the pretrained parameters and only train prompts. Though the computational cost is significant reduced compared with fine tuning, the computation for the model inference still remains the same. 
Future works may further consider the synergy between the task of  parameter-efficient training and model compression (\emph{e.g.} distillation or pruning), so that the resulting model can also benefit from reduced computation cost during inference.

\ifdefined\isarxiv

\else 
\bibliography{ref}
\bibliographystyle{plain}

\fi


\appendix
\section*{Appendix}

{\bf Roadmap.} 
In Section~\ref{sec:preli}, we provide a number of basic notations. In Section~\ref{sec:def}, we provide several basic definitions and discuss some related work about previous theoretical softmax regression results. In Section~\ref{sec:theory}, we provide a complete proof for our major theoretical result in this paper. We present our final result in Section~\ref{sec:main_theory}.

\section{Preliminaries}\label{sec:preli}

 For any positive integer $n$, we use $[n]$ to denote set $\{1,2,\cdots,n\}$. For any function $f$, we use $\wt{O}(g)$ to denote $g \cdot \poly(\log g)$.

\paragraph{Vector}

For a length-$n$ vector $z$, we use $\exp(z)$ to denote a length-$n$ vector that its $i$-th entry is $\exp(z_i)$. 

For a length-$n$ vector $z$, we use $\| z \|_2$ to represent its $\ell_2$ norm, i.e., $\| z \|_2 := ( \sum_{i=1}^n x_i^2 )^{1/2}$. For a length-$n$ vector $z$, we use $\| z \|_{\infty}$ to denote $\max_{i \in [n]} |z_i|$.

For a length-$n$ vector $z \in \R^n$, we use $\diag(z)$ to generate a diagonal matrix where each entry on the ($i,i$)-diagonal is $z_i$ for every $i \in [n]$.

We use ${\bf 1}_n$ to represent a length-$n$ vector where all the coordinates are ones. Similarly, we use ${\bf 0}_n$ to represent a length-$n$ vector where all the values are zeros.

\paragraph{PSD}

We say $W \succeq Z$ (positive semi-definite) if $x^\top  W x \geq x^\top Z x$ for all vector $x$.

\paragraph{Matrix Related}

For an $n$ by $d$ matrix $C$, we use $\nnz(C)$ to denote the number of non-zero entries of $C$, i.e., $\nnz(C) := | \{ (i,j) \in [n] \times [d] ~|~ C_{i,j} \neq 0 \} |$

For a diagonal matrix $D \in \R^{n \times n}$, we say $D$ is a $k$-sparse diagonal matrix, i.e., $k = |\{ i \in [n] ~|~ D_{i,i} \neq 0 \}|$.

For any matrix $Z \in \R^{n \times k}$, we denote the spectral norm of $Z$ by $\| Z \|$, i.e., 
\begin{align*}
\| Z \| := \sup_{x\in\R^k} \frac{ \| Z x \|_2 }{ \| x \|_2 }.
\end{align*}

For a matrix $Q$, we use $\sigma_{\max}(Q)$ to denote the largest singular value of $Q$. We use $\sigma_{\min}(Q)$ to denote the smallest singular value of $Q$.

\paragraph{Matrix Computation}

We use $\omega$ to denote the exponent of matrix multiplication, i.e., $n^{\omega}$ denotes the time of multiplying an $n \times n$ matrix with another $n \times n$ matrix. Currently $\omega \approx 2.373$ \cite{w12,lg14,aw21}.

\paragraph{Calculus Related}
We use $\circ$ notation by following the literature's \cite{lsz23,dls23,gsy23,lsx+23,ssz23}.
Suppose that we're given two column vectors $x, y \in \R^n$, we use $x \circ y$ to denote a column vector that $(x\circ y)_i$ is $x_iy_i$.
\section{Related Work about Theoretical Attention Regression Results}\label{sec:def}

 In this paper, we focus on the direction of regression tasks \cite{gms23,lsz23,dls23,lsx+23,ssz23}. The goal of this section will review the linear regression (Definition~\ref{def:linear_regression}), exponential regression (Definition~\ref{def:exp_regression}), rescaled softmax regression (Definition~\ref{def:rescaled_softmax_regression}), softmax regression (Definition~\ref{def:exp_regression}).

\begin{definition}[Linear regression]\label{def:linear_regression}
Given a matrix $A \in \R^{n \times d}$ and $b \in \R^n$, the goal is to solve  
\begin{align*}
\min_{x \in \R^d} \| A x - b \|_2.
\end{align*}

For convenient, let us $u(x)$ to denote $\exp(Ax)$.
 
\end{definition}

\begin{definition}[Exponential Regression, see \cite{gms23,lsz23}]\label{def:exp_regression}
Suppose we are given a length $n$ vector $b$, and an $n$ by $d$ size matrix $A$, our goal is to optimize
\begin{align*}
\min_{x \in \R^d} \| u(x) - b \|_2.
\end{align*}
\end{definition}

\begin{definition}[Rescaled Softmax Regression, see \cite{gsy23}]\label{def:rescaled_softmax_regression}
Suppose we are given a length $n$ vector $b$, and an $n$ by $d$ size matrix $A$, our goal is to optimize
\begin{align*}
\min_{x \in \R^d} \| u(x) - \langle u(x) , {\bf 1}_n \rangle \cdot b \|_2
\end{align*}
\end{definition}
 
\begin{definition}[Softmax Regression, see \cite{dls23,lsx+23,ssz23}]\label{def:softmax_regression}
Suppose we are given a length $n$ vector $b$, and an $n$ by $d$ size matrix $A$, our goal is to optimize
\begin{align*}
\min_{x \in \R^d} \| \langle u(x), {\bf 1}_n \rangle^{-1} \cdot u(x) - b \|_2.
\end{align*}
\end{definition}

\section{Theoretical Guarantees}\label{sec:theory}

In Section~\ref{sec:theory:def}, we provide several basic definitions. In Section~\ref{sec:theory:gradient_f}, we explain how to compute the gradient of function $f$. In Section~\ref{sec:theory:gradient_log_f}, we show how to compute the gradient of function $\log f(x)$. In Section~\ref{sec:theory:hessian_log_f}, we explain how to compute the Hessian of function $\log f(x)$. In Section~\ref{sec:theory:hessian_log_f_b}, we compute the hessian of inner product between $\log f(x)$ and $b$. In Section~\ref{sec:theory:hessian_cent}, we compute the Hessian of cross entropy loss function. In Section~\ref{sec:theory:hessian_pd}, we show Hessian is positive definite. In Section~\ref{sec:theory:hessian_lipschitz}, we prove that Hessian is Lipschitz.

\subsection{Function Definition}
\label{sec:theory:def}

We define
\begin{definition}\label{def:u}
We define $u(x)$ as follows
\begin{itemize}
    \item $u(x) = \exp(Ax)$ 
\end{itemize}
\end{definition}

\begin{definition}\label{def:v}
We define $v(x)$ as follows
\begin{itemize}
    \item $v(x) = \exp(Ax)$  
\end{itemize}
\end{definition}
Previous \cite{lsz23} studies three types of hyperbolic functions $\exp( \cdot )$, $\cosh( \cdot )$ and $\sinh( \cdot )$. We mainly focus on $\exp(\cdot)$ function.

\begin{definition}[Normalized coefficients, Definition 5.4 in \cite{dls23}]
\label{def:alpha}
    We define $\alpha : \R^d \rightarrow \R$ as follows
    \begin{align*}
    \alpha(x) := \langle u(x), {\bf 1}_n \rangle.
    \end{align*}
\end{definition}

We define function softmax $f$ as follows
\begin{definition}[Function $f$, Definition 5.1 in \cite{dls23}]\label{def:f}
Suppose that we're given an $n \times d$ matrix $A$. Let ${\bf 1}_n$ denote a length-$n$ vector that all the coordinates are ones.  
We define prediction function $f: \R^d \rightarrow \R^n$ as follows 
\begin{align*}
f(x) := \langle u(x) , {\bf 1}_n \rangle^{-1} \cdot u(x) .
\end{align*}
\end{definition}

\begin{fact}\label{fac:f}
Let $f(x)$ be defined as Definition~\ref{def:f}. Then we have
\begin{itemize}
    \item Part 1. $\langle f(x), {\bf 1}_n \rangle = 1$
    \item Part 2. $\| f(x) \|_1 = 1$
    \item Part 3. $\| f(x) \|_2 \leq 1$
\end{itemize}
\end{fact}
\begin{proof}
The proof is straightforward. For more details, we refer the readers to \cite{dls23}.
\end{proof}

We define the $\ell_2$ loss
\begin{definition}\label{def:L_exp}
We define
\begin{align*}
    L_{\exp} := 0.5\| f(x) - b \|_2^2. 
\end{align*}    
\end{definition}

Previous work \cite{ssz23} only considers entropy, here we consider cross entropy instead.
\begin{definition}[Cross Entropy]\label{def:cent}
We define $L_{\cent} : \R^d \rightarrow \R$, 
\begin{align*}
L_{\cent}(x) :=  - \langle b, \log f(x) \rangle
\end{align*}
\end{definition}

\begin{definition}\label{def:L_reg}
Suppose we're given an $n \times d$ matrix $A$ and $W = \diag (w) \in \R^{n \times n} $ where $w \in \R^n$ is a vector, we define $L_{\reg} : \R^d \rightarrow \R$
\begin{align*}
L_{\reg} (x) := 0.5 \| W A x\|_2^2    
\end{align*} 
\end{definition}

\subsection{Gradient Computation for Function \texorpdfstring{$f$}{}}\label{sec:theory:gradient_f}

We present a calculus tool from previous work \cite{dls23} (for example, we refer the readers to Lemma 5.6 in \cite{dls23}).
\begin{lemma}[]\label{lem:gradient_f}
If the following conditions hold
\begin{itemize}
    \item Given matrix $A \in \R^{n \times d}$ and a vector $b \in \R^n$.
    \item Suppose that function $\alpha : \R^d \rightarrow \R$ be defined in Definition~\ref{def:alpha}.
    \item Suppose that function $f : \R^d \rightarrow \R^n$ be defined in Definition~\ref{def:f}.
\end{itemize}
 For each $i \in [d]$, we have
 \begin{itemize}
    \item Part 1. 
    \begin{align*}
        \frac{\d f (x) }{\d x_i}  =  & ~ -  \langle f(x), A_{*,i}\rangle \cdot f(x) + f(x) \circ A_{*,i}
    \end{align*} 
    \item Part 2.  
    \begin{align*}
        \langle \frac{\d f(x) }{\d x_i} , A_{*,i} \rangle = - \langle f(x), A_{*,i} \rangle^2 + \langle f(x), A_{*,i} \circ A_{*,i} \rangle
    \end{align*}
    \item Part 3.  
    \begin{align*}
        \langle \frac{\d f(x) }{\d x_i} , A_{*,j} \rangle = - \langle f(x), A_{*,i} \rangle \cdot \langle f(x), A_{*,j} \rangle + \langle f(x), A_{*,i} \circ A_{*,j} \rangle
    \end{align*}
\end{itemize}
\end{lemma}

\subsection{Gradient Computation for Function \texorpdfstring{$\log f(x)$}{}}\label{sec:theory:gradient_log_f}

In this section, we explain how to compute the gradient of $\log f(x)$.
\begin{lemma}\label{lem:gradient_log_f}
If the following condition holds
\begin{itemize}
    \item Suppose that function $f$ is defined in Definition~\ref{def:f}.
\end{itemize}
We have
\begin{itemize}
\item Part 1.
\begin{align*}
    \frac{\d \log f(x)}{ \d x_i} =  - \langle f(x) , A_{*,i} \rangle \cdot {\bf 1}_n + A_{*,i}
\end{align*}
\item Part 2.
\begin{align*}
 \langle \frac{\d \log f(x)}{ \d x_i} , b \rangle = \langle A_{*,i} , b \rangle- \langle f(x), A_{*,i} \rangle \cdot \langle b, {\bf 1}_n \rangle 
\end{align*}
\item Part 3.
\begin{align*}
    \frac{\d}{\d x_i} L_{\cent}(x) = \langle f(x), A_{*,i} \rangle \cdot \langle b, {\bf 1}_n \rangle - \langle A_{*,i}, b \rangle
\end{align*}
\end{itemize}
\end{lemma}
\begin{proof}

{\bf Proof of Part 1.}

For all index $j \in [n]$, we can compute the gradient with respect to $x_i$
\begin{align*}
\frac{\d \log f(x)_j}{ \d x_i} = f(x)_j^{-1} \frac{\d f(x)_j}{ \d x_i}
\end{align*}
Then we group the $n$ coordinates, we get
\begin{align*}
\frac{\d \log f(x)}{ \d x_i} 
= & ~ f(x)^{-1} \circ \frac{\d f(x)}{ \d x_i} \\
= & ~ f(x)^{-1} \circ ( -  \langle f(x), A_{*,i}\rangle f(x) + f(x) \circ A_{*,i} ) \\
= & ~ -\langle f(x), A_{*,i} \rangle f(x)^{-1} \circ f(x) + f(x)^{-1} \circ f(x) \circ A_{*,i} \\
= & ~ - \langle f(x) , A_{*,i} \rangle \cdot {\bf 1}_n + A_{*,i}
\end{align*}

{\bf Proof of Part 2.}
We have 
\begin{align*}
     \langle \frac{\d \log f(x)}{ \d x_i} , b \rangle = & ~ \langle - \langle f(x) , A_{*,i} \rangle \cdot {\bf 1}_n + A_{*,i}, b \rangle \\
     = & ~\langle A_{*,i} , b \rangle- \langle f(x), A_{*,i} \rangle \cdot \langle b, {\bf 1}_n \rangle ,
\end{align*}
where the first step follows from Part 1 and the second step follows from simple algebra.

{\bf Proof of Part 3.}
The proof directly follows from Part 2 and Definition of $L_{\cent}(x)$ (See Definition~\ref{def:cent}). 
\end{proof}

\subsection{Hessian Computation for Function \texorpdfstring{$\log f(x)$}{}}\label{sec:theory:hessian_log_f}

In this section, we will show how to compute the Hessian for function $\log f(x)$. 
\begin{lemma}\label{lem:hessian_log_f}
If the following conditions hold
\begin{itemize}
    \item Let $f$ be defined as Definition~\ref{def:f}.
\end{itemize}
Then we have
\begin{itemize}
    \item Part 1.
    \begin{align*}
        \frac{\d^2 \log f(x)}{\d x_i^2} = ( \langle f(x), A_{*,i} \rangle^2 - \langle f(x), A_{*,i} \circ A_{*,i} \rangle ) \cdot {\bf 1}_n
    \end{align*}
    \item Part 2.
    \begin{align*}
        \frac{\d^2 \log f(x)}{\d x_i \d x_j} = ( \langle f(x), A_{*,i} \rangle \langle f(x), A_{*,j} \rangle - \langle f(x), A_{*,i} \circ A_{*,j} \rangle ) \cdot {\bf 1}_n
    \end{align*}
\end{itemize}
\end{lemma}
\begin{proof}

{\bf Proof of Part 1.}

We have
\begin{align*}
\frac{\d^2 \log f(x) }{\d x_i^2}
= & ~ \frac{\d }{\d x_i} ( \frac{\d \log f(x)}{\d x_i} ) \\
= & ~ \frac{\d }{\d x_i} ( - \langle f(x), A_{*,i} \rangle \cdot {\bf 1}_n + A_{*,i} ) \\
= & ~ -  \frac{\d }{\d x_i} ( \langle f(x), A_{*,i} \rangle ) \cdot {\bf 1}_n \\
= & ~ ( \langle f(x), A_{*,i} \rangle^2 - \langle f(x), A_{*,i} \circ A_{*,i} \rangle ) \cdot {\bf 1}_n
\end{align*}
where the 2nd step comes from Part 1 of Lemma~\ref{lem:gradient_log_f}, the 3rd step follows from $A_{*,i}$ is independent of $x$, and the forth step follows from Part 2 of Lemma~\ref{lem:gradient_f}.

{\bf Proof of Part 2.}

Similarly, we can provide a proof for Part 2.
\end{proof}

\subsection{Hessian Computation for Function \texorpdfstring{$\langle \log f(x), b \rangle $}{}}\label{sec:theory:hessian_log_f_b}
The goal of this section is to prove Lemma~\ref{lem:hessian_log_f_b}.
\begin{lemma}\label{lem:hessian_log_f_b}
If the following conditions hold
\begin{itemize}
    \item Let $f$ be defined as Definition~\ref{def:f}.
\end{itemize}
Then we have
\begin{itemize}
    \item Part 1.
    \begin{align*}
        \langle \frac{\d^2 \log f(x)}{\d x_i^2} , b \rangle = ( \langle f(x), A_{*,i} \rangle^2 - \langle f(x), A_{*,i} \circ A_{*,i} \rangle ) \cdot \langle {\bf 1}_n, b \rangle
    \end{align*}
    \item Part 2.
    \begin{align*}
       \langle \frac{\d^2 \log f(x)}{\d x_i \d x_j} , b \rangle = ( \langle f(x), A_{*,i} \rangle \langle f(x), A_{*,j} \rangle - \langle f(x), A_{*,i} \circ A_{*,j} \rangle ) \cdot \langle {\bf 1}_n ,b \rangle
    \end{align*}
\end{itemize}
\end{lemma}
\begin{proof}
The proof directly follows from Lemma~\ref{lem:hessian_log_f}.
\end{proof}

\subsection{Hessian Computation for Function \texorpdfstring{$L_{\cent}(x)$}{}}\label{sec:theory:hessian_cent}

For convenient of analyzing the $d \times d$ Hessian matrix, we will start with defining $n \times n$ matrix $B$.
\begin{definition}\label{def:B}
We define $B(x) \in \R^{n \times n}$ as follows
\begin{align*}
    B(x):= \langle {\bf 1}_n, b \rangle \cdot ( \diag(f(x) ) - f(x) f(x)^\top )
\end{align*}
\end{definition}

\begin{lemma}\label{lem:hessian_cent}
If the following conditions hold
\begin{itemize}
    \item Let $f$ be defined as Definition~\ref{def:f}.
    \item Let $L_{\cent}$ be defined as Definition~\ref{def:cent}
    \item Let $B$ be defined as Definition~\ref{def:B}
\end{itemize}
Then we have
\begin{itemize}
    \item Part 1.
    \begin{align*}
       \frac{\d^2}{ \d x_i^2} L_{\cent} = ( - \langle f(x), A_{*,i} \rangle^2 + \langle f(x), A_{*,i} \circ A_{*,i} \rangle ) \cdot \langle {\bf 1}_n, b \rangle
    \end{align*}
    \item Part 2.
    \begin{align*}
       \frac{\d^2}{ \d x_i \d x_j} L_{\cent}= ( - \langle f(x), A_{*,i} \rangle \langle f(x), A_{*,j} \rangle + \langle f(x), A_{*,i} \circ A_{*,j} \rangle ) \cdot \langle {\bf 1}_n ,b \rangle
    \end{align*}
    \item Part 3.
    \begin{align*}
        \frac{\d^2 }{\d x^2} L_{\cent} = A^\top B(x) A
    \end{align*}    
\end{itemize}
\end{lemma}
\begin{proof}
The proof trivially follows from Lemma~\ref{lem:hessian_log_f_b} and Definition~\ref{def:B}.
\end{proof}

\subsection{Hessian is Positive Definite}\label{sec:theory:hessian_pd}

Previous work \cite{dls23} doesn't consider cross entropy into the final loss function. Here we generalize previous lemma so that cross entropy is also being considered. 
\begin{lemma}[A cross entropy generalization of Lemma 6.3 in \cite{dls23}]\label{lem:hessian_psd}
Suppose the following conditions hold
\begin{itemize}
    \item Let $A \in \R^{n \times d}$, $R \geq 4$, $l > 0$, suppose that $R_0 = \exp( O(R^2 + \log n) ) $ 
    \item \begin{align*}
    L(x)= \underbrace{ L_{\reg}(x) }_{ \mathrm{Definition~\ref{def:L_reg}} } + \underbrace{ L_{\cent}(x) }_{ \mathrm{Definition~\ref{def:cent}} } + \underbrace{ L_{\exp}(x) }_{ \mathrm{Definition~\ref{def:L_exp}}  }.
    \end{align*}
    \item Let $\wt{B}(x) = B(x) +W^2$ 
\end{itemize}
Then we have
\begin{itemize}
    \item Part~1. $\min_{i\in [n]} w_i^2 \geq 10 R_0 + l/\sigma_{\min} (A)^2$, then we have 
    \begin{align*}
        \frac{\d^2 L}{ \d x^2} \succeq l \cdot I_d
    \end{align*}
    \item Part~2. $\min_{i\in [n]} w_i^2 \geq 10^4 \cdot R_0 + l/\sigma_{\min} (A)^2$, then we have
   \begin{align*}
        (1-0.01)\cdot \wt{B}(x) \preceq W^2 \preceq (1-0.01)\cdot \wt{B}(x).
    \end{align*}
\end{itemize}
\end{lemma}
\begin{proof}
Using the definition of $B$ for $L_{\cent}$(see Definition~\ref{def:B}), definition/bound of $B$ for $L_{\exp}$ (see \cite{dls23}), and tools developed in Section 6 in \cite{dls23}, we can finish the proof.
\end{proof}

\subsection{Hessian is Lipschitz}\label{sec:theory:hessian_lipschitz}
Previous work \cite{dls23} doesn't consider cross entropy into the final loss function. Here we generalize previous lemma so that cross entropy is also being considered. 

\begin{lemma}[A cross entropy version of Lemma 7.1 in \cite{dls23}]\label{lem:hessian_lipschitz}
Suppose the following condition holds
\begin{itemize}
    \item Let $H(x) = \frac{ \d^2 L}{\d x^2} $  and $R > 4$
    \item Suppose that $\max\{ \| x \|_2 , \| y \|_2 \} \leq R$, and $\max\{ \| A \|, \| b\|_2 \} \leq R$
    \item $\| A (x-y) \|_\infty < 0.01$
\end{itemize}
Then we have 
\begin{align*}
    \| H(x) - H(y)\| \leq  n^{4} \exp (O( R^2 + \log n) ) \cdot \| x - y \|_2
\end{align*}
\end{lemma}
\begin{proof}
Using the definition of $B$ for $L_{\cent}$(see Definition~\ref{def:B}), definition/bound of $B$ for $L_{\exp}$ (see \cite{dls23}), and tools developed in Section 7 in \cite{dls23}, we can finish the proof.
\end{proof}

\begin{algorithm}[!ht]\caption{Our Algorithm.}\label{alg:main}
\begin{algorithmic}[1]
\Procedure{OurAlgorithm}{$A \in \R^{n \times d},b \in \R^n,w \in \R^n, \epsilon, \delta$} \Comment{Theorem~\ref{thm:main:formal}} 
    \State We choose $x_0$ 
    \State $T \gets \log( \| x_0 - x^* \|_2 / \epsilon )$ \Comment{ $T$ denotes the number of iterations}  
    \For{$t=0 \to T$}  
        \State Implicitly formulate exact Hessian and use that to construct an approximate Hessian $\wt{H}$ (similar as Section 8 in \cite{dls23})
        \State Compute gradient 
        \State $\wt{H} \gets A^\top \wt{D} A$ 
        \State $x_{t+1} \gets x_t + \wt{H}^{-1} g$  
    \EndFor
    \State $\wt{x}\gets x_{T+1}$
    \State \Return $\wt{x}$
\EndProcedure
\end{algorithmic}
\end{algorithm}

\section{Main Theoretical Guarantees}\label{sec:main_theory}

Previous work \cite{dls23} has proved the similar result without considering the cross entropy. We generalize the techniques in previous paper \cite{dls23} from only considering $\ell_2$ task loss to considering both $\ell_2$ task loss and cross entropy loss ($L_{\cent}$ see formal definition in Definition~\ref{def:cent}). Our algorithm is a version of approximate Newton method, such methods have been widely used in many optimization tasks \cite{cls19,lsz19,b20,jkl+20,sy21,jswz21,dly21,gs22,syyz22,hjs+22,qszz23,dls23}. In this work, we focus on the approximate Newton method along the line of \cite{syyz22,dls23}.

\begin{theorem}[Formal version of Theorem~\ref{thm:main:informal}]\label{thm:main:formal}

Let $x^*$ denote an length-$d$ vector that is satisfying,
\begin{align*}
    \arg \min_{x \in \R^d} \underbrace{ L_{\exp} }_{ \mathrm{Definition~\ref{def:L_exp}} } + \underbrace{ L_{\cent} }_{ \mathrm{Definition~\ref{def:cent}} } + \underbrace{ L_{\reg} }_{ \mathrm{Definition~\ref{def:L_reg}} }
\end{align*}
Suppose the following conditions are holding:
\begin{itemize}
    \item $R \geq 4$, $g(x^*) = {\bf 0}_d$.
    \item $\| x^* \|_2 \leq R$, $\| A \| \leq R$, $\| b \|_2  \leq R$.
    \item $M = \exp(O( R^2 + \log n) )$.
    \item $\min_{i \in [n]}w_{i}^2 \geq 100 M + l/\sigma_{\min}(A)^2$   
    \item Suppose that $\epsilon \in (0,0.1)$ is the final and $\delta \in (0,0.1)$ is the failure probability.
    \item Suppose $x_0$ satisfy condition $M \| x_0 - x^* \|_2 \leq 0.1 l$.
    \item Suppose that $T = \log(\| x_0 - x^* \|_2/ \epsilon)$ 
\end{itemize}
Then there is a randomized algorithm (Algorithm~\ref{alg:main}) such that
\begin{itemize}
    \item it runs $T$ iterations
    \item in each iteration, it spends time\footnote{Here $\omega$ denotes the exponent of matrix multiplication. Currently $\omega \approx 2.373$.}
    \begin{align*}
            O( (\nnz(A) + d^{\omega} ) \cdot \poly(\log(n/\delta)). 
    \end{align*}
    \item generates a vector $\wt{x} \in \R^d$ that is satisfying 
    \begin{align*}
        \| \wt{x} - x^* \|_2 \leq \epsilon
    \end{align*}
    \item the succeed probability is $1-\delta$
\end{itemize}

\end{theorem}
\begin{proof}
The high level framework of our theorem is similar to previous work about exponential regression \cite{lsz23}, softmax regression \cite{dls23} and rescaled softmax regression \cite{gsy23}. Similarly as previous work \cite{lsz23,dls23,gsy23,ssz23}, we use the approximate newton algorithm (for example see Section 8 in \cite{dls23}). So in the proof, we only focus on the difference about the Hessian positive definite lower bound and Hessian Lispchitz property.

Using Lemma~\ref{lem:hessian_psd} and Lemma~\ref{lem:hessian_lipschitz} and approximate Newton algorithm analysis in \cite{dls23}, then we complete the proof.
\end{proof}

\ifdefined\isarxiv
\bibliography{ref}
\bibliographystyle{alpha}
\else

\fi

\end{document}